\documentclass[runningheads]{llncs}
\usepackage{graphicx}
\usepackage{soul}
\usepackage{hyperref}
\hypersetup{
    colorlinks=true,
    }
\usepackage{tikz}
\usepackage{comment}
\usepackage{amsmath,amssymb} 
\usepackage{color}
\usepackage{booktabs}
\usepackage{multirow}
\usepackage{pifont}
\usepackage{bm}  
\newcommand{\eg}{\textit{e.g.}}
\newcommand{\ie}{\textit{i.e.}}
\usepackage[accsupp]{axessibility}  
\newcommand{\cmark}{\ding{51}}%
\newcommand{\xmark}{\ding{55}}%

\newcommand{\modelname}{Zero-Shot Temporal Action Detection via Vision-Language Prompting}
\newcommand{\shortmodelname}{STALE}

\begin{document}
\pagestyle{headings}
\mainmatter
\def\ECCVSubNumber{100}  

\title{Zero-Shot Temporal Action Detection via Vision-Language Prompting} 


\titlerunning{Zero-Shot TAD via Vision-Language Prompting}
\author{Sauradip Nag\inst{1,2} \and
Xiatian Zhu\inst{1,3} \and
Yi-Zhe Song\inst{1,2} \and
Tao Xiang\inst{1,2}}
\authorrunning{Nag et al.}
\institute{CVSSP, University of Surrey, UK \and
iFlyTek-Surrey Joint Research Centre on Artificial Intelligence, UK \and
Surrey Institute for People-Centred Artificial Intelligence, University of Surrey, UK\\
\email{\{s.nag,xiatian.zhu,y.song,t.xiang\}@surrey.ac.uk}}
\maketitle

\begin{abstract}
Existing temporal action detection (TAD) methods rely on large training data including segment-level annotations, limited to recognizing previously seen classes alone during inference.
Collecting and annotating a large training set for each class of interest is costly and hence unscalable.
Zero-shot TAD (ZS-TAD) resolves this obstacle by 
enabling a pre-trained model to recognize any unseen action classes.
Meanwhile, ZS-TAD is also much more challenging with significantly less investigation. 
Inspired by the success of zero-shot image classification 
aided by vision-language (ViL) models such as CLIP,
we aim to tackle the more complex TAD task.
An intuitive method is to integrate an off-the-shelf proposal detector with CLIP style classification.
However, due to the {\em sequential} localization (\eg, proposal generation) and classification design, it is prone to localization error propagation.
To overcome this problem, in this paper we propose a novel {\em zero-\underline{S}hot \underline{T}emporal \underline{A}ction detection model via Vision-\underline{L}anguag\underline{E} prompting} ({\shortmodelname}). 
Such a novel design effectively eliminates the dependence between localization and classification by breaking the route for error propagation in-between.
We further introduce an interaction mechanism between classification and localization for improved optimization.
Extensive experiments on standard ZS-TAD video benchmarks show that our {\shortmodelname} significantly outperforms state-of-the-art alternatives.
Besides, our model also yields superior results on supervised TAD over recent strong competitors. The PyTorch implementation of {\shortmodelname} is available on \href{https://github.com/sauradip/STALE}{https://github.com/sauradip/STALE}.

\end{abstract}

\section{Introduction}

The introduction of large pretrained Visual-Language (ViL) models (\eg, CLIP \cite{radford2021learning} and ALIGN \cite{jia2021scaling}) has surged recently the research attempts
on {\em zero-shot transfer} to diverse downstream tasks
via {\em prompting}.
Central in this research line is a favourable ability of {\em synthesizing 
the classification weights} from natural language which can describe almost all the classes of interest,
along with a strong image encoder trained with an objective of aligning a massive number of potentially noisy image-text pairs in a common feature space.

Whilst existing works mostly focus on image recognition tasks \cite{zhou2021learning,gao2021clip}, how to capitalize the knowledge of such ViL models for natural video understanding tasks (\eg, temporal action detection \cite{lin2019bmn,xu2020g,caba2015activitynet,nag2022gsm,nag2021temporal,nag2022pftm,nag2021few,xu2021boundary,xu2021boundary,xu2020g}) remains unsolved largely.
There are several challenges to be solved.
{\bf\em First}, as those {\em public accessible} large ViL models
are often pretrained with big image data, temporal structured information is absent which is limited for video representation.
{\bf\em Second}, ViL models favor classification type of tasks
{\em without built-in detection capability} in formulation.
That is, their optimization assumes implicitly limited background 
{\em w.r.t} the foreground content with the training data. In contrast,
natural videos often come with a large, variable proportion of background, and detecting the content of interest (\eg, human actions) is critical \cite{alwassel2018diagnosing}.
It is non-trivial to overcome these obstacles.
On using ViL models for temporal action detection (TAD), indeed a recent attempt \cite{ju2021prompting} is made.
Specifically, \cite{ju2021prompting} presents a two-stage video model:
generating many action proposals with an off-the-shelf pre-trained proposal detector (\eg, BMN \cite{lin2019bmn}), followed by proposal classification.
This method has obvious limitations:
(1) It is unable to adapt the localization module (\ie, proposal detection) which is pretrained and frozen throughout.
(2) As a result, the compatibility between the localization and classification is limited, further leading to a localization error propagation problem due to the sequential localization and classification pipeline.

\begin{figure}[t]
\centering
    \includegraphics[scale=0.35]{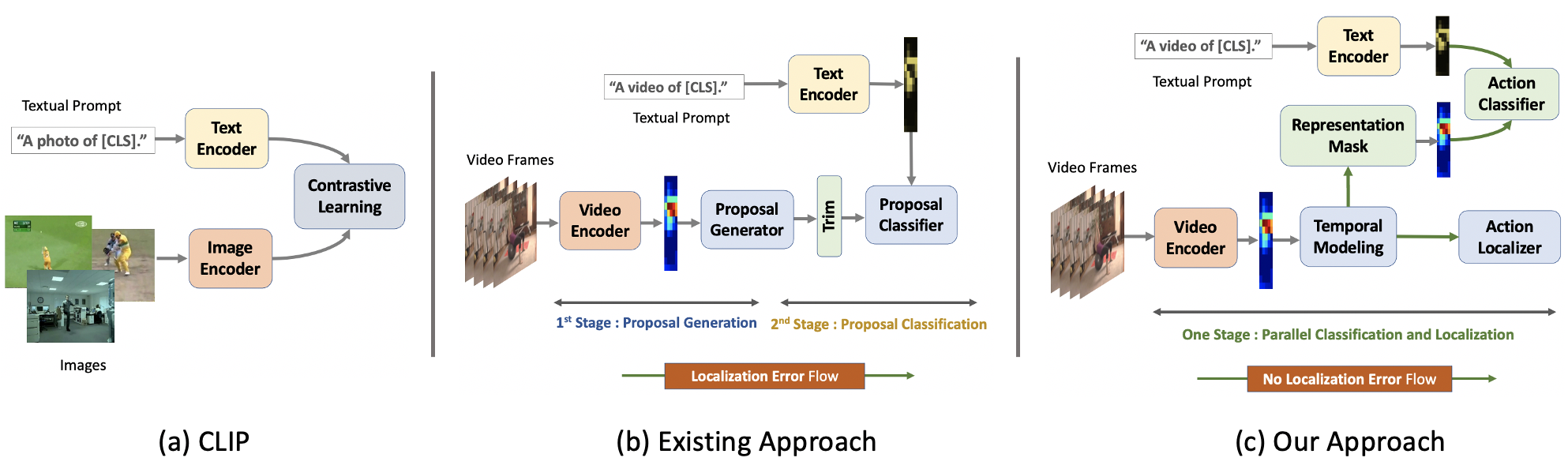}
    \caption{
    Illustration of the (a) standard vision-language model CLIP \cite{radford2021learning}.
    (b) Existing ZS-TAD method \cite{ju2021prompting} suffers from the intrinsic localization error propagation problem, due to the {\em sequential} localization (\eg, proposal generation) and classification design.
    (c) We solve this problem by designing a {\shortmodelname} model with a {\em parallel} localization and classification architecture.
   }
    \vspace{-.2in}
    \label{fig:intropic}
\end{figure}

In this paper, to overcome the aforementioned limitations, we formulate a novel one-stage zero-shot temporal action localization architecture
characterized with {\em parallel} classification and localization.
This parallel head design naturally eliminates error propagation.
Importantly, we introduce a learnable localization module based on 
a notion of {\em representation masking} -- which is class-agnostic
and hence generalizable to unseen classes.
It can be optimized end-to-end, together with the classification component
powered by a ViL model for zero-shot transfer, consequently solving the component compatibility problem.
For improving cross-modal (from image to video) task adaptation,
we further design an inter-stream alignment regularization 
along with the self-attention mechanism.
We term our method {\bf\em  Zero-\underline{S}hot \underline{T}emporal \underline{A}ction Detection Model via Vision-\underline{L}anguag\underline{E} Prompting} (\shortmodelname).

\noindent\textbf{Contributions.} 
\textbf{(1)} We investigate the under-studied yet critical problem of how to capitalize large pretrained ViL models for zero-shot temporal action localization (ZS-TAD) in untrimmed videos.
\textbf{(2)} We present a novel one-stage model, {\shortmodelname},
featured with a parallel classification and localization design
interleaved by a learnable class-agnostic representation masking component
for enabling zero-shot transfer to unseen classes.  
For enhancing cross-modal task adaptation,
we introduce an inter-stream alignment regularization 
in the Transformer framework.
\textbf{(3)} Extensive experiments on standard ZS-TAD video benchmarks 
show that our {\shortmodelname} outperforms state-of-the-art alternative methods, often by a large margin.
Besides, our model can be also applied to the fully supervised TAD setting and achieve superior performance in comparison to recent supervised competitors.

\section{Related Works}
\textbf{Vision-language models} 
There have been a series of works
on the interaction of computer vision and natural language
processing fields, \eg, text-to-image retrieval \cite{wang2019camp}, image
caption \cite{xu2015show}, visual question answering \cite{antol2015vqa}, and so on. Among these works, vision language (ViL) pre-training has attracted growing attention during
the past few years \cite{lei2021less,lu2019vilbert}. 
As a milestone, Radford et
al. \cite{radford2021learning} devise a large-scale pretraining ViL model, named CLIP,
trained with a contrastive learning strategy on 400 million image-text pairs.
It shows impressive zero-shot transferable ability over 30 classification datasets. 
Since then, many follow-ups have been proposed, including improved
training strategy (\eg, CoOp \cite{zhou2021learning}, CLIP-Adapter \cite{gao2021clip},
Tip-adapter \cite{zhang2021tip}). 
In video
domains, similar idea has also been explored for transferable representation learning \cite{miech2020end}, text based action localization \cite{paul2021text}. 
CLIP has also been used very recently in action recognition (\eg, ActionCLIP \cite{wang2021actionclip}) and TAD \cite{ju2021prompting}. 
Conceptually, all these methods use CLIP in a classification perspective.
Commonly a two-stage pipeline is applied, where the first stage needs to crop the foreground followed by aligning the foreground using CLIP in the second stage. 
However, this strategy often suffers from the error propagation problem,
that is, the errors from first stage would flow into the second stage and potentially amplify. 
In this work, we tackle this challenge by designing a one-stage ZS-TAD architecture featured with parallel classification and localization.

\noindent \textbf{Temporal action detection}
\textcolor{black}{Substantial progress has been made in TAD. Inspired by object detection in static images \cite{ren2016faster},
R-C3D \cite{xu2017r} uses anchor boxes by following the design of {proposal generation and classification}.
With a similar model design, TURN \cite{gao2017turn} aggregates local features to represent snippet-level features for temporal boundary regression and classification. SSN \cite{zhao2017temporal} decomposes an action instance into three stages (starting, course, and ending)
and employs structured temporal pyramid pooling
to generate proposals.
BSN \cite{lin2018bsn} predicts the start, end and actionness at each temporal location and generates proposals with high start and end probabilities. The actionness was further improved in BMN \cite{lin2019bmn} via
additionally generating a boundary-matching confidence map for improved proposal generation. 
GTAN \cite{long2019gaussian}
improves the proposal feature pooling procedure with a learnable Gaussian kernel for weighted averaging. G-TAD \cite{xu2020g}
learns semantic and temporal context via graph convolutional networks for more accurate proposal generation. BSN++ \cite{su2020bsn++} further extends BMN with a complementary boundary generator to capture rich context.
CSA \cite{sridhar2021class} enriches the proposal temporal context via attention transfer. Recently, VSGN \cite{zhao2021video} improves short-action localization using a cross-scale multi-level pyramidal architecture. Commonly, the existing TAD models mostly 
adopt a 2-stage {\em sequential}
localization and classification architecture.
This would cause the localization error propagation problem
particularly in low-data setting such as ZS-TAD.
Our {\shortmodelname} is designed to address this limitation by designing a single stage model, thereby removing
the dependence between localization and classification
and cutting off the error propagation path.
}

\noindent\textbf{Zero-shot temporal action detection}
Zero-shot learning (ZSL) is designed to recognize new classes that are not seen during training \cite{xian2018zero}. The idea is to learn shared knowledge from prior information and then transfer that knowledge from seen classes
to unseen classes \cite{niu2018zero,qin2017zero}. 
Visual attributes (\eg, color, shape, and any properties) are the typical forms of prior information. 
For example, Lampert et al. \cite{lampert2013attribute} prelearned the attribute classifiers independently to accomplish
ZSL on unseen classes, while Parikh et al. \cite{parikh2011relative} learned relative attributes.
Despite promising results on ZSL, attribute-based
methods have poor scalability because the attributes need to be manually defined. 
Semantic embeddings of seen and unseen
concepts, another type of prior information, can solve
this scalability problem \cite{xu2015semantic}. They are generally learned in an unsupervised manner such as Word2Vec \cite{goldberg2014word2vec} or
GloVe \cite{pennington2014glove}. 
Zhang et al. \cite{zhang2020zstad} firstly applied zero-shot learning on TAD using Word2Vec. 
Very recently, EffPrompt \cite{ju2021prompting} used image-text pretraining from CLIP \cite{radford2021learning} for ZS-TAD.
However, due to a two-stage design, this method also has the error propagation problem, in addition to its inability of learning the action localization module.
We address all these limitations by introducing a new one-stage ZS-TAD architecture.

\section{Methodology}
We aim to efficiently steer an image-based ViL model (CLIP \cite{radford2021learning}) to 
tackle dense video downstream tasks such as 
Zero-Shot Temporal Action Detection (ZS-TAD) in untrimmed videos.
This is essentially a model adaptation process with the aim to leverage the rich semantic knowledge from large language corpus.

\subsection{Preliminaries: Visual-Language Pre-training}
The key capability of CLIP is to align the embedding spaces of visual and language data \cite{radford2021learning}.
It consists of two encoders, namely an image encoder (\eg, ResNet \cite{he2016deep} or ViT \cite{dosovitskiy2020image}) and a text encoder (\eg, Transformer \cite{vaswani2017attention}). To learn rich transferable semantic knowledge, CLIP leverages 400 million image-text pairs during training. To exploit CLIP's knowledge for downstream classification tasks, an effective approach is to construct a set of text prompts with a template such as \textit{``a photo of }$[CLS]\textit{."}$, where $[CLS]$ can be replaced by any class name of interest. Given an image, one can then use CLIP to compute the similarity scores between this image and the text prompts in the embedding space and take the class with the highest score as the prediction. Recently, a couple of works \cite{zhou2021learning,gao2021clip} have demonstrated that CLIP can obtain strong classification performance with few or even zero training examples per class. We raise an interesting question: {\bf \em Whether
the impressive ability of CLIP can be transferred to more complex vision tasks like dense prediction?} 

This extended transfer could be intrinsically nontrivial. {\em Firstly}, how to leverage the visual-language pre-trained model in dense prediction tasks is a barely studied problem especially in zero-shot setup \cite{rao2021denseclip}. 
A simple method is to only use the image encoder of CLIP.
However, we argue that the language priors with
the pretrained text encoder are also of great importance
and should be leveraged together. 
{\em Secondly}, 
transferring the
knowledge from CLIP to dense prediction is more difficult than classification tasks,
due to the substantial task discrepancy involved.
The pretraining focuses on 
global representation learning of both images and texts,
which is incompatible for local pixel-level outputs
as required in downstream tasks.
RegionCLIP \cite{zhong2021regionclip} recently solves this problem in a 2-stage design including class-agnostic masking generalizable to unseen classes, followed by CLIP style classification. EffPrompt \cite{ju2021prompting} similarly tackles a dense video understanding task TAD. Nonetheless, this approach suffers from the notorious localization error propagation challenge. 
\begin{figure*}[h]
\begin{center}
  \includegraphics[scale=0.48]{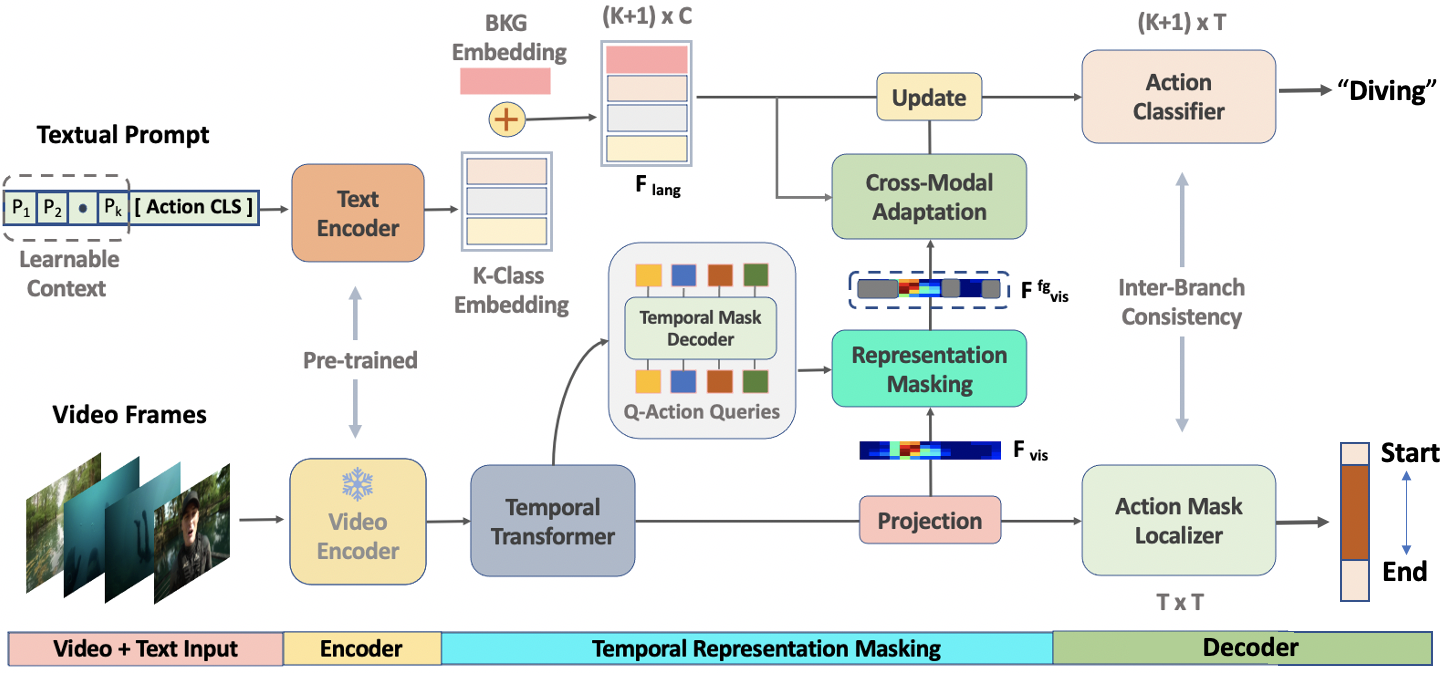}
\end{center}
\caption{\textbf{Overview of the proposed {\em \modelname} (\shortmodelname) method.} 
Given an untrimmed video $V$, (a) we first extract a sequence of $T$ snippet features with a pre-trained frozen video encoder and conduct self-attention learning using temporal embedding to obtain the snippet embedding $E$ with a global context.
(b) For each snippet embedding, we then predict a classification score $P$ with the classification stream by masking the foreground feature and aligning with the text encoder embedding to obtain a classifier output $P$. The other branch of snippet embedding is used by action mask classifier to obtain a foreground mask $M$ in parallel,
(c) both of which are further used for consistency refinement
at the feature level.
}
\label{fig:network}
\end{figure*}

\subsection{Language Guided Temporal Action Detection}

To solve the aforementioned issues, we propose a language guided proposal-free framework. It can better leverage the language-priors of the pre-trained CLIP model. In the following sections, we start by describing the problem scenario and notations followed by introducing
the idea for model adaptation through prompt learning. We then
discuss how masking can help maintain the zero-shot transfer property in a one-stage design. Lastly, we discuss about the refinement of boundaries.

\subsubsection{Problem definition}
We assume a dataset $D$ with the training $D_{train} = \{V_{i},\psi_{i}\}_{i=1}^N$ and validation $D_{val}$ sets.
Each untrimmed training video $V_i$ is labeled with temporal segmentation 
$\Psi_i = \{(\psi_j, \xi_j, y_j)\}_{j=1}^{M_i}$
where $\psi_{j}$/$\xi_{j}$ denote the start/end time, $y_j$ is the action category, and $M_i$ is the action instance number. 
Respectively, $y_{j}$ refers to one of the training ($D_{train}$) action labels in the text format for recognition, \eg, $y_{j}$ replaces the $[CLS]$ token in a sentence of $``$a photo of $[CLS]"$. 
We consider both closed-set and open-set scenarios.
In the closed-set scenario, the action categories for training and evaluation are identical, \ie, $D_{train} = D_{val}$. While in the open-set case, the action categories for training and evaluation are disjoint, \ie, $D_{train} \cap D_{val} = \phi$.

\subsubsection{Visual language embedding }
Given a varying length untrimmed video $V$, following the standard practice \cite{xu2020g,lin2019bmn} we first sample \textcolor{black}{a very large} $T$ equidistantly distributed temporal snippets (points) over the entire length.

\noindent \textit{Visual embedding}: \textcolor{black}{To extract features from the video snippets we use a frozen pre-trained video encoder (\eg, a I3D \cite{carreira2017quo}, CLIP \cite{radford2021learning}) to extract RGB $X_{r} \in \mathbb{R}^{d \times T}$ and optical flow features $X_{o} \in \mathbb{R}^{d \times T}$ at the snippet level, where $d$ denotes the feature dimension. We then concatenate them as $E=[ X_{r};X_{o}] \in \mathbb{R}^{2d \times T}$. Each snippet is a short sequence of (\eg, 16 in our case) consecutive frames.}
While $F$ contains local spatio-temporal information, it lacks
a global context critical for TAD. We hence leverage the self-attention mechanism \cite{vaswani2017attention} to learn the global context. Formally, we set the input (\texttt{query}, \texttt{key}, \texttt{value}) of a multi-head Transformer encoder $\mathcal{T}()$ as the features ({$F$,$F$,$F$}) (Fig. \ref{fig:network})). Positional encoding is not applied as it is found to be detrimental (see the \texttt{supplementary material}). The final video snippet embedding is then obtained as 
\begin{equation}
    F_{vis} = \mathcal{T}(E) \in \mathbb{R}^{C \times T},
\end{equation}
with $C$ being the embedding dimension. 

\noindent\textit{Textual encoding}: 
For textual embedding, we use a standard CLIP pre-trained Transformer \cite{radford2021learning} with learnable prompt, similar to \cite{zhou2021learning}, and as opposed to handcrafted prompt used in CLIP \cite{radford2021learning}. By making it learnable, textual contexts can now achieve better transferability in downstream classification tasks by directly optimizing the contexts using back-propagation. The input to the text encoder, formed as:
\begin{equation}
    F_{lan} = [G_{p};G^{k}_{e}] \hfill where \hfill G_{p} \in \mathbb{R}^{\hat{N} \times C^{'}},
\end{equation}
are the learnable textual contexts, with $\hat{N}$ as the length of the contexts. 
The embedding $G^{k}_{e} \in \mathbb{R}^{C^{'}}$ represents the textual embedding from CLIP vocabulary for each of the name of $k$-th class. The textual embedding for background class is not directly available from CLIP vocabulary, but necessary in TAD.
To solve this issue, we learn a specific background embedding, denoted as $G^{bg}_{e} \in \mathbb{R}^{C^{'}}$. We append this to the action class embedding $G^{k}_{e}$, making $F_{lan}$ an embedding with $K+1$ classes. The background embedding $G^{bg}_{e}$ is initialized randomly. 

\subsubsection{Class agnostic representation masking}
We introduce a novel {\em class agnostic representation masking} concept 
for enabling the usage of a ViL model for ZS-TAD.
This is conceptually inspired by mask-transformer \cite{cheng2021per} with focus on a totally different problem (image segmentation {\em without} the aid of ViL model).
Specifically, given the snippet embedding $F_{vis}$ per video and $N_z$ mask queries particularly introduced in this work, we leverage a transformer decoder \cite{carion2020end}
to generate $N_z$ latent embeddings. 
Then we pass each latent embedding through 
a masking projection layer to obtain a mask embedding 
for each segment as $B_{q} \in \mathbb{R}^{q \times C}$ where $q$ indexes a query.  
A binary mask prediction {\em w.r.t} each query can be then calculated as:
\begin{equation}
 L_{q} = \sigma(B_{q}*F_{vis}) \in \mathbb{R}^{q \times T},
\end{equation}
where $\sigma$ is sigmoid activation. 
As such, each snippet location is associated with $q$ queries. To choose the optimal query per location, we deploy a tiny MLP to weigh these queries in a location specific manner. This is realized by learning a weight vector $W_q \in \mathbb{R}^{1 \times q}$  as: 
\begin{equation}
    \hat{L} = \sigma(W_{q}*L_{q} + b_{q}) \in \mathbb{R}^{T}.
\end{equation}
We then binarize this mask at a threshold $\theta_{bin}$ and select the foreground mask, denoted by $\hat{L}_{bin}$. 
To obtain the foreground features $F^{fg}_{vis}$, we use $\hat{L}_{bin}$ to retrieve 
the snippet embedding $F_{vis}$.
Given the binary nature of this foreground feature mask $\hat{L}_{bin}$,
our representation masking can be first optimized on seen action classes
and further generalize to unseen classes.

\subsubsection{Vision-language cross-modal adaption}
Intuitively, integrating the descriptions of visual contexts is likely to enrich the text representation. For example, $``${a video of a man \textit{playing kickball} in a big park}$”$ is richer than $``${a video of a man \textit{playing kickball}}$"$. This motivates us to investigate how to use visual contexts to refine the text features. 
%
Specifically, we leverage the contextual-level visual feature to guide the text feature to adaptively explore informative regions of a video. For cross-attention, we adopt the standard architecture of the transformer \cite{vaswani2017attention}.
Concretely, the cross attention module consists of a self-attention layer, a co-attention layer and a feed forward network formed as:
\begin{equation}
    \hat{E}_{c} = \mathcal{T}_{c}(F_{lan},F^{fg}_{vis}, F^{fg}_{vis}),
\end{equation}
where $\mathcal{T}_{c}$ is a transformer layer, taking $F_{lan}$ as the \texttt{query}, and $F^{fg}_{vis}$ as the \texttt{key} and \texttt{value}. This module encourages the text features to find most related visual clues across the foreground snippets. 
We then update the text features through a residual connection:
\begin{equation}
    \hat{F}_{lan} = F_{lan} + \alpha \hat{E}_{c},
\end{equation}
where $\alpha \in \mathbb{R}^{C}$ is a learnable parameter to control the scaling of the residual. $\alpha$ is initialized with very small values (\eg, $10^{-3}$) to maximally preserve the language priors of text features.

\subsubsection{Parallel classification and mask prediction}
Our TAD head is featured with parallel classification and mask prediction as detailed below.

\noindent{\bf\em (I) Contextualized vision-language classifier}:
In the standard training process of CLIP \cite{radford2021learning}, the global feature is normally used during contrastive alignment. In general, it estimates the snippet-text score pairs by taking the average pooling of snippet features and then uses it with the language features. However, this formulation is unsuitable for dense classification tasks like TAD where each temporal snippet needs to be assigned with a class label. 
Under this consideration, we instead use the updated textual features $\hat{F}_{lan} \in \mathbb{R}^{(K+1) \times C}$ and the masked foreground feature $F^{fg}_{vis} \in \mathbb{R}^{T \times C}$ as follows:
\begin{equation}
\label{Eq:7}
    \mathcal{P} = \hat{F}_{lan}*(F^{fg}_{vis})^{T},
\end{equation}
where $\mathcal{P} \in \mathbb{R}^{(K+1) \times T}$ represents the classification output, where each snippet location $t \in T$ is assigned with a probability distribution $p_{t} \in \mathbb{R}^{(K+1) \times 1}$. Note, 
$l_{2}$ normalized along the channel dimension is applied prior to Eq. \eqref{Eq:7}.

\noindent{\bf\em (II) Action mask localizer}: 
In parallel to the classification stream, this stream
predicts 1-D masks of action instances across the whole temporal span
of the video. Since the 1-D masks are conditioned on the temporal location $t$, we exploit dynamic convolution \cite{chen2020dynamic}. 
This is because, in contrast to standard convolution, dynamic filters allow to leverage separate network branches to generate a filter at each snippet location. As a consequence, the dynamic filters can learn the context of the action (background) instances at each snippet location individually. More specifically, given the $t$-th snippet $F_{vis}(t)$, it outputs a 1-D mask vector $m_{t} = [q_{1},...,q_{T}] \in \mathbb{R}^{T \times 1}$ with each element $q_{i} \in [0,1] (i \in [1,T])$ indicating foreground probability of $i$-th snippet. This is implemented by a stack of three 1-D dynamic convolution layers $H_{m}$ as follows:
\begin{equation}
    \mathcal{M} = sigmoid(H_{m}(F_{vis})),
\end{equation}
\noindent where $t$-th column of $\mathcal{M}$ is the temporal mask prediction by $t$-th snippet. More details on dynamic convolution formulation is given in \texttt{supplementary material}.

\subsection{Model Training and Inference}
\noindent{\bf Label assignment} To train our one-stage {\shortmodelname}, the ground-truth needs to be arranged into the designed format. Concretely, given a training video with temporal intervals and class labels, we label all the snippets of a single action instance with the same action class. All the snippets off from action intervals
are labeled as background. 
For an action snippet of a particular instance in the class stream, we assign the video-length binary instance mask at the same snippet location in the action mask stream.
Each mask is action instance specific.
All snippets of a specific action instance share
the same mask. Refer to \texttt{supplementary material} for more details.

\noindent{\bf Learning objective}
The classification stream is composed of a simple cross-entropy loss. For a training snippet, we denote $y \in \mathbb{R}^{(K+1) \times T}$ the ground-truth class label, and $p \in \mathcal{P}$  the classification output. We compute the classification loss as:
\begin{equation}
    L_{c} = CrossEntropy(p,y).
\end{equation}
For the segmentation mask branch, we combine a weighted cross entropy and binary dice loss \cite{milletari2016v}.
Formally, for a snippet location, 
we denote $\bm{m} \in \mathbb{R}^{T\times 1}$ the predicted segmentation mask, and
$\bm{g} \in \mathbb{R}^{T\times 1}$ the ground-truth mask.
The loss for the segmentation mask branch is formulated as:
\begin{equation}
\centering \small
\begin{aligned}
    L_m = \beta_{fg}
    \sum_{t=1}^{T}
    \bm{g}(t) \log(\bm{m}(t)) + 
    \beta_{bg}
    \sum_{t=1}^{T} (1-\bm{g}(t)) \log(1-\bm{m}(t)) 
    \\
    +
    \lambda_2
    \Big( 1 - 
    \frac{\bm{m}^\top \bm{g}}
    {\sum_{t=1}^{T} \big(\bm{m}(t)^2 + \bm{g}(t)^2 \big)} 
    \Big),
\end{aligned}
\end{equation}
where $\beta_{fg}$/$\beta_{bg}$
is the inverse of foreground/background snippet's proportion.
We set the loss trade-off coefficient $\lambda_2 = 0.4$.

We further impose a 1-D action completeness loss formed by binary cross entropy (BCE). It can penalize the foreground masking output $\hat{L} \in \mathbb{R}^{T \times 1}$.
Given a ground truth one-hot foreground mask $\hat{g} \in \mathbb{R}^{T \times 1}$, we design the loss to model the completeness of foreground as follows:
\begin{equation}
{L}_{comp} = 
-\Big ( \hat{g}*\log(\hat{L}) + \log(\hat{g})*\hat{L} \Big ).
\label{eq:LR}
\end{equation}

\noindent{\em Inter-branch consistency}
In our \shortmodelname, there is structural consistency {in terms of foreground} between the class and mask labels by design. 
To leverage this consistency for improved optimization, we formulate 
the consistency loss as:
\begin{equation}
     L_{const} = 1 - \texttt{cosine}\Big( 
    \hat{F}_{clf},
    \hat{F}_{mask}
    \Big),
\label{eq:clmask}
\end{equation}
where $\hat{F}_{clf} = topk(argmax((P_{bin}*E_{p})[:K,:]))$ is the features obtained from the top scoring foreground snippets obtained from the thresholded classification output $P_{bin} := \eta(P - \theta_{c})$
with $\theta_{c}$ the threshold and $E_{p}$ obtained by passing the embedding $E$ into a 1D conv layer for matching the dimension of $P$.
The top scoring features from the mask output $M$ are obtained similarly as:
$\hat{F}_{mask} = topk(\sigma(1DPool(E_{m}*M_{bin})))$
where $M_{bin} := \eta(M - \theta_{m})$ is a binarization of mask-prediction $M$, $E_{m}$ is obtained by passing the embedding $E$ into a 1D conv layer for matching the dimension of $M$, and $\sigma$ is sigmoid activation.

\noindent{\bf \em Overall objective} The overall objective loss function of our \shortmodelname{} is defined as:
$L = L_{c} + L_{m} + L_{comp} + L_{const}$.  
This loss is used to train the model end-to-end, whilst leaving out the pre-trained video encoder frozen due to the GPU memory constraint.

\subsubsection{Model inference}
\textcolor{black}{
At test time, we generate action instance predictions for each test video by the classification $\bm{P}$ and mask $\bm{M}$ predictions.
For $\bm{P}$, we only consider the snippets whose class probabilities are greater than $\theta_{c}$ and select top scoring snippets.
For each such top scoring action snippet,
we then obtain the
temporal mask by thresholding the $t_i$-th column of $\bm{M}$ using the localization threshold ${\Theta}$. To produce sufficient candidates,
we use a set of thresholds $\Theta=\{\theta_i\}$. 
For each candidate, we compute a confidence score $s$
by multiplying the classification and max-mask scores.
SoftNMS \cite{bodla2017soft} is finally applied to obtain top scoring results.
}

\section{Experiments}
\noindent\textbf{Datasets}
We conduct extensive experiments on two popular TAD benchmarks.
(1) ActivityNet-v1.3 \cite{caba2015activitynet} has 19,994 videos from 200 action classes. We follow the
standard setting to split all videos into training, validation and testing subsets
in ratio of 2:1:1. (2) THUMOS14 \cite{idrees2017thumos} has 200 validation videos and 213 testing
videos from 20 categories with labeled temporal boundary and action class

\noindent\textbf{Implementation details}
For fair comparison with existing TAD works we use Kinetics pre-trained I3D \cite{carreira2017quo} as the video encoder for both ActivityNet and THUMOS. We also use the two-stream features as used in \cite{ju2021prompting} for fair-comparison. For comparing with CLIP based TAD baselines, we also adopt the image and text encoders from pre-trained CLIP (ViT-B/16+Transformer). For model-adaptation, the visual encoder is kept frozen, the trainable parts are textual prompt embeddings, text encoder, temporal embedding module, temporal masking modules and TAD decoder heads. For the CLIP encoders, the video frames are pre-processed to $224 \times 224$ spatial resolution, and the maximum number of textual tokens is $77$ (following the original CLIP design). Each video’s feature sequence F is rescaled to T = 100/256 snippets for AcitivtyNet/THUMOS using linear interpolation. Our model is trained for 15 epochs using Adam with learning rate of {$10^{-4}/10^{-5}$ for ActivityNet/THUMOS respectively.}

\subsection{Comparative Results}
\subsubsection{Zero-shot action detection}

\noindent\textbf{Setting}
In this section, we evaluate on the open-set scenario where $D_{train } \cap D_{val} = \phi$, \ie, action categories for training and testing are disjoint. 
We follow the setting and dataset splits proposed by \cite{ju2021prompting}. More specifically, we initiate two evaluation settings on THUMOS14 and ActivityNet1.3:
(1) Training with 75\% action categories and testing on the left 25\% action categories; 
(2) Training with 50\% categories and testing on the left 50\%
categories. To ensure statistical significance, we conduct 10
random samplings to split categories for each setting, following \cite{ju2021prompting}.

\noindent\textbf{Competitors} 
As there is no open-source implementation for \cite{ju2021prompting}, we use their same reported baselines.
More specifically,
(1) One baseline using BMN \cite{lin2019bmn} as proposal generator and CLIP \cite{radford2021learning} with hand crafted prompt. This is the same baseline as reported in \cite{ju2021prompting}. This serves as a 2-stage TAD baseline using CLIP. We term it \texttt{B-I}.
(2) One comparative CLIP based TAD EffPrompt \cite{ju2021prompting}.
(3) One CLIP + TAD model for one-stage baseline: As ZS-TAD is a relatively new problem, we need to implement the competitors by extending existing TAD methods using CLIP by ourselves. 
We select a variant of CLIP for dense prediction task \cite{rao2021denseclip} using the existing CLIP pre-trained weights. We term this baseline, DenseCLIP (w/ CLIP Pretrained Image Encoder) + TAD, as \texttt{B-II}. 
The text-encoders however are identical for both the baselines using CLIP pre-training weights. However, we could not compare with the earlier zero-shot TAD method ZS-TAD \cite{zhang2020zstad} due to unavailability of code and no common data-split between \cite{zhang2020zstad} and \cite{ju2021prompting}. 

\noindent\textbf{Performance} The ZS-TAD results are reported in Table~\ref{tab:zshot}. With 50\% labeled data, our \shortmodelname{} surpasses both 1-stage and 2-stage baselines, as well as the CLIP based TAD method \cite{ju2021prompting} by a good margin on ActivityNet. This suggests that our representation masking is able to perform the zero-shot transfer better than conventional 2-stage designs. This also indicates that the localization-error propagation hurts in low training data regime. Also noted that the one-stage baseline (\texttt{B-II}) performs worst among all other baselines. This suggests that a class-agnostic masking is necessary to achieve strong performance in ZS-TAD. The performance of our model however drops on THUMOS14 on stricter metrics, possibly due to 
the mask decoder suffering from foreground imbalance.
We observe similar trend with $75\%$ labeled data, our approach is again superior than all other competitors on both datasets. It is also interesting to note that the  one-stage baseline (\texttt{B-II}) has a larger performance gap with the two-stage baseline (\texttt{B-I}), around $3.1\%$ avg mAP on ActivityNet. This however reduces to $0.6\%$ when the number of labeled data increase, suggesting that the one-stage baseline has potential to improve with more data.

\subsubsection{Closed-set action detection}
\noindent\textbf{Setting} Closed-set action detection refers to the common setting, where the model is trained and evaluated on videos of the same action categories, \ie, $D_{train} = D_{val}$. For a fair comparison, we use the same dataset splits as in the literature.

\noindent\textbf{Competitors} We considered the following methods for comparison. (1) Seven representative TAD methods with I3D encoder backbone;
(2) One CLIP+TAD method EffPrompt \cite{ju2021prompting}; (3) One two-stage CLIP based TAD baseline \texttt{B-I}; (4) One single-stage CLIP based TAD baseline \texttt{B-II};  (5) We also created another single-stage baseline by replacing CLIP-Pretrained encoder with Kinetics pre-trained Video-Encoder (\eg, I3D). We term it \texttt{B-III}.

\noindent\textbf{Performance} From the results of Table \ref{tab:sup}, we observe that with more labeled data, our approach also surpasses existing TAD approaches often by a large margin. This is consistent over both the datasets. Thus the text-embedding is indeed helpful for our design. We also observe that our method performs better by almost similar margin when we use different feature backbone (\eg, CLIP). Thus it proves our design is feature agnostic. Another key observation is that our single-stage baseline (\texttt{B-II}) performs significantly better than \cite{ju2021prompting} by at least $5\%$ in avg mAP for both the datasets. Thus justifies that our parallel design is better alternative for CLIP based approaches.

\begin{table}[t]
\centering
\setlength{\tabcolsep}{3pt}
\caption{\textbf{Results of Zero-Shot Action Detection}}
\label{tab:zshot}
\begin{tabular}{@{}c|c|cccccc|cccc@{}}
\toprule

\multirow{2}{*}{Train Split} &
  \multirow{2}{*}{Methods} &
  \multicolumn{6}{c|}{Thumos 14} &
  \multicolumn{4}{c}{ActivityNet v1.3} \\ \cmidrule(l){3-12} 
 &
   &
  0.3 &
  0.4 &
  0.5 &
  0.6 &
  \multicolumn{1}{c|}{0.7} &
  Avg &
  0.5 &
  0.75 &
  \multicolumn{1}{c|}{0.95} &
  Avg \\ \midrule
\multirow{4}{*}{\begin{tabular}[c]{@{}c@{}}75\% Seen \\   25\% Unseen\end{tabular}} &
   \texttt{B-II} &
  28.5 &
  20.3 &
  17.1 &
  10.5 &
  \multicolumn{1}{c|}{6.9} &
  16.6 &
  32.6 &
  18.5 &
  \multicolumn{1}{c|}{5.8} &
  19.6 \\
 &
  \texttt{B-I} &
  33.0 &
  25.5 &
  18.3 &
  11.6 &
  \multicolumn{1}{c|}{5.7} &
  18.8 &
  35.6 &
  20.4 &
  \multicolumn{1}{c|}{2.1} &
  20.2  \\
 &
  EffPrompt  &
  39.7 &
  31.6 &
  23.0 &
  14.9 &
  \multicolumn{1}{c|}{7.5} &
  23.3 &
  37.6 &
  22.9 &
  \multicolumn{1}{c|}{3.8} &
  23.1
  \\\cmidrule(l){2-12} 
 &
  \textbf{\shortmodelname} &
  \textbf{40.5} &
  \textbf{32.3} &
  \textbf{23.5} &
  \textbf{15.3} &
  \multicolumn{1}{c|}{\textbf{7.6}} &
  \textbf{23.8} &
  \textbf{38.2} &
  \textbf{25.2} &
  \multicolumn{1}{c|}{\textbf{6.0}} &
  \textbf{24.9} \\ \midrule
\multirow{4}{*}{\begin{tabular}[c]{@{}c@{}}50\% Seen\\ 50\% Unseen\end{tabular}} &
  \texttt{B-II} &
  21.0 &
  16.4 &
  11.2 &
  6.3 &
  \multicolumn{1}{c|}{3.2} &
  11.6 &
  25.3 &
  13.0 &
  \multicolumn{1}{c|}{3.7} &
  12.9 \\ &
  \texttt{B-I} &
  27.2 &
  21.3 &
  15.3 &
  9.7 &
  \multicolumn{1}{c|}{4.8} &
  15.7 &
  28.0 &
  16.4 &
  \multicolumn{1}{c|}{1.2} &
  16.0 \\
  &
  EffPrompt  &
  37.2 &
  29.6 &
  \textbf{21.6} &
  \textbf{14.0} &
  \multicolumn{1}{c|}{\textbf{7.2}} &
  21.9 &
  32.0 &
  19.3 &
  \multicolumn{1}{c|}{2.9} &
  19.6 \\
\cmidrule(l){2-12} 
 &
  \textbf{\shortmodelname} &
  \textbf{38.3} &
  \textbf{30.7} &
  21.2 &
  13.8 &
  \multicolumn{1}{c|}{7.0} &
  \textbf{22.2} &
  \textbf{32.1} &
  \textbf{20.7} &
  \multicolumn{1}{c|}{\textbf{5.9}} &
  \textbf{20.5} \\ \bottomrule
\end{tabular}
\end{table}

\begin{table}[h]
\caption{\textbf{Comparison with state-of-the-art on closed-set setting}}
\label{tab:sup}
\resizebox{\textwidth}{!}{
\footnotesize
\begin{tabular}{@{}c|c|c|cccccc|cccc@{}}
\toprule
\multirow{2}{*}{Methods} &
  \multirow{2}{*}{Mode} &
  \multirow{2}{*}{\begin{tabular}[c]{@{}c@{}}Encoder\\ Backbone\end{tabular}} &
  \multicolumn{6}{c|}{THUMOS14} &
  \multicolumn{4}{c}{ActivityNet v1.3} \\ \cmidrule(l){4-13} 
            &          &            & 0.3  & 0.4  & 0.5  & 0.6  & \multicolumn{1}{c|}{0.7}  & Avg  & 0.5  & 0.75 & \multicolumn{1}{c|}{0.95} & Avg  \\ \midrule
TALNet      & RGB+Flow & I3D        & 53.2 & 48.5 & 42.8 & 33.8 & \multicolumn{1}{c|}{20.8} & 39.8 & 38.2 & 18.3 & \multicolumn{1}{c|}{1.3}  & 20.2 \\
GTAN        & RGB+Flow & P3D        & 57.8 & 47.2 & 38.8 & -    & \multicolumn{1}{c|}{-}    & -    & 52.6 & 34.1 & \multicolumn{1}{c|}{8.9}  & 34.3 
\\
MUSES       & RGB+Flow & I3D        & 68.9 & 64.0 & 56.9 & 46.3 & \multicolumn{1}{c|}{31.0} & 53.4 & 50.0 & 34.9 & \multicolumn{1}{c|}{6.5}  & 34.0 \\
VSGN        & RGB+Flow & I3D        & 66.7 & 60.4 & 52.4 & 41.0 & \multicolumn{1}{c|}{30.4} & 50.1 & 52.3 & 36.0 & \multicolumn{1}{c|}{8.3}  & 35.0 \\
Context-Loc & RGB+Flow & I3D        & 68.3 & 63.8 & 54.3 & 41.8 & \multicolumn{1}{c|}{26.2} & -    & 56.0 & 35.2 & \multicolumn{1}{c|}{3.5}  & 34.2 \\
BU-TAL & RGB+Flow & I3D & 53.9 & 50.7 & 45.4 & 38.0 & \multicolumn{1}{c|}{28.5} & 43.3 & 43.5 & 33.9 & \multicolumn{1}{c|}{9.2} &30.1 \\
\texttt{B-III}     & RGB+Flow & I3D        & 68.3 & 62.3 & 51.9 & 38.8 & \multicolumn{1}{c|}{23.7} & -    & 47.2 & 30.7 & \multicolumn{1}{c|}{8.6}  & 30.8 \\
\midrule
\textbf{\shortmodelname} &
  RGB+Flow &
  I3D &
  \textbf{68.9} &
  \textbf{64.1} &
  \textbf{57.1} &
  \textbf{46.7} &
  \multicolumn{1}{c|}{\textbf{31.2}} &
  \textbf{52.9} &
  \textbf{56.5} &
  \textbf{36.7} &
  \multicolumn{1}{c|}{\textbf{9.5}} &
  \textbf{36.4} 
  \\ \midrule
TALNet      & RGB      & I3D        & 42.6 & –    & 31.9 & –    & \multicolumn{1}{c|}{14.2} & –    & –    & –    & \multicolumn{1}{c|}{–}    & –    \\
A2Net       & RGB      & I3D        & 45.0 & 40.5 & 31.3 & 19.9 & \multicolumn{1}{c|}{10.0} & 29.3 & 39.6 & 25.7 & \multicolumn{1}{c|}{2.8}  & 24.8 \\

\texttt{B-I}   & RGB      & CLIP &  36.3  & 31.9  & 25.4 & 17.8 & \multicolumn{1}{c|}{10.4 } & 24.3  & 28.2  & 18.3  & \multicolumn{1}{c|}{3.7 }  & 18.2 \\
\texttt{B-II}     & RGB & CLIP        & 57.1  & 49.1  & 40.4  & 31.2  & \multicolumn{1}{c|}{23.1 } & 40.2    & 51.5 & 33.3 & \multicolumn{1}{c|}{6.6}  & 32.7 \\ 
EffPrompt   & RGB      & CLIP & 50.8 & 44.1 & 35.8 & 25.7 & \multicolumn{1}{c|}{15.7} & 34.5 & 44.0 & 27.0 & \multicolumn{1}{c|}{5.1}  & 27.3 \\ \midrule
\textbf{\shortmodelname} &
  RGB &
  CLIP &
  \textbf{60.6 } &
  \textbf{53.2} &
  \textbf{44.6} &
  \textbf{36.8} &
  \multicolumn{1}{c|}{\textbf{26.7}} &
  \textbf{44.4} &
  \textbf{54.3} &
  \textbf{34.0} &
  \multicolumn{1}{c|}{\textbf{7.7}} &
  \textbf{34.3} \\ \bottomrule
\end{tabular}
}
\end{table}

\subsection{Ablation}

\noindent{\bf Localization error propagation analysis }
To examine the effect of localization error propagation
with previous TAD models,
we design a proof-of-concept experiment 
by measuring the performance drop between ground-truth proposals
and pseudo proposals.
 Due to unavaiability of training code in \cite{ju2021prompting}, we carefully re-created \texttt{B-I} following the details in \cite{ju2021prompting}.
For our {\shortmodelname} model, we contrast ground-truth and output masks.
This experiment is tested on ActivityNet with $75\%$ label split.
Table \ref{tab:err} shows that the proposal based
TAD baseline suffers almost double performance degradation from localization (\ie, proposal) error due to its sequential localization and classification design.
This verifies the advantage of {\shortmodelname}'s parallel design.

\noindent{\bf Necessity of representation masking}
\textcolor{black}{To validate the role of {\em representation masking} in generalization of the classifier, we perform experiments in the $75\%$ split setting.
Firstly, we compare our approach by removing the Maskformer-Decoder \cite{cheng2021per} from the pipeline of {\shortmodelname} and passing the temporal feature $F_{vis}$ directly into the cross-modal adaptation module for classification. As shown in Table \ref{tab:featmask}, we observe a sharp drop of $14\%$ in avg mAP, justifying that foreground features are indeed necessary to align the text embedding. This problem is also profound in DenseCLIP baseline (\texttt{B-II}) as illustrated in Table \ref{tab:zshot}. 
We also tested another alternative to masking, \eg, vanilla 1-D CNN. It is evident that Maskformer-Decoder benefits from learning the query embedding over vanilla 1-D CNN. We also observe that increasing the number of queries improves the overall localization performance but comes with high memory footprint. Further, due to memory constraints of query-matching step of the decoder output during classification, we learn a MLP to reduce the cost. We verify that class-agnostic representation-masking has much stronger generalization in localizing unseen action categories. Interestingly, we observe that the maximum {\shortmodelname} can achieve avg mAP of $35.8\%$  indicating a room for further improvement.}

\begin{table}[t]
    \begin{minipage}{.45\linewidth}
    
      \caption{Analysis of localization error propagation on ActivityNet with 75\% split. 
GT: Ground-Truth.}
      \centering
      \resizebox{.99\textwidth}{!}{
      \setlength{\tabcolsep}{8pt}
        \begin{tabular}{@{}cll@{}}

\toprule
\multirow{2}{*}{Metric}                             & \multicolumn{2}{c}{mAP}       \\ \cmidrule(l){2-3} 
                                                     & 0.5           & Avg           \\ \midrule
\multicolumn{3}{c}{Baseline-I(\texttt{B-I})}                                                          \\ \midrule
\multicolumn{1}{c|}{GT proposals}        & 56.2          & 47.1          \\ 
\multicolumn{1}{c|}{Predicted proposals} & 34.8         & 19.9          \\\midrule
\multicolumn{3}{c}{\bf \shortmodelname}                                                          \\ \midrule
\multicolumn{1}{c|}{GT masks}
& 53.6          & 42.3          \\ 
\multicolumn{1}{c|}{{Predicted masks}}           & {38.2}  & {24.9}  \\

\bottomrule
\end{tabular}
\label{tab:err}
        }
    \end{minipage}%
    \hfill
    \begin{minipage}{.5\linewidth}
      \centering
        \caption{Analysis of Representation-Masking on 75\% seen split on ActivityNet}
\label{tab:featmask}
\begin{tabular}{@{}c|c|cc@{}}
\toprule
\multirow{2}{*}{Masking Method}     & \multirow{2}{*}{\# Queries} & \multicolumn{2}{c}{mAP}                            \\ \cmidrule(l){3-4} 
                                    &                          & \multicolumn{1}{c|}{0.5}           & Avg           \\ \midrule
No Mask                             & -                        & \multicolumn{1}{c|}{21.7}          & 10.9          \\
GT-Mask                             & -                        & \multicolumn{1}{c|}{54.3}          & 35.8          \\ \midrule
1-D CNN                             & -                        & \multicolumn{1}{c|}{33.5}          & 20.1          \\ \midrule
\multirow{3}{*}{Maskformer \cite{cheng2021per}} & 5                        & \multicolumn{1}{c|}{37.5}          & 24.4          \\
                                    & \textbf{20}              & \multicolumn{1}{c|}{\textbf{38.2}} & \textbf{24.9} \\
                                    & 100                      & \multicolumn{1}{c|}{39.0}          & 25.1          \\ \bottomrule
\end{tabular}
    \end{minipage} 
\end{table}


\noindent\textbf{Importance of text encoder}
\textcolor{black}{
We ablate the effect of text encoder fine-tuning. 
Note the video encoder is frozen due to memory constraint.
We use CLIP \cite{radford2021learning} pre-trained text transformer (denoted as CLIP-text) in this experiment. 
It can be observed from Table \ref{tab:text} that using a text-encoder is indeed important for the overall performance.
Further, fine-tuning the text encoder is also effective due to the
large domain gap between CLIP pre-training and TAD task.
}

\noindent\textbf{Effect of temporal modeling}
\textcolor{black}{Recall that we use a multi-head Transformer (w/o positional encoding) for temporal modeling  in {\shortmodelname}. 
We evaluate this design choice by comparing
(I) a 1D CNN with 3 dilation rates 
(1, 3, 5) each with 2 layers,
 and 
(II) a multi-scale Temporal Convolutional Network MS-TCN \cite{farha2019ms}.
Each CNN design substitutes the default Transformer while remaining all the others.
{Table \ref{tab:emb} shows that 
the Transformer is clearly superior to both CNN alternatives. 
This suggests that our default design captures stronger contextual learning capability even in low-data setting like ZS-TAD. 
}}
\begin{table}[!htb]
    \begin{minipage}{.5\linewidth}
      \centering
        \caption{Importance of {\em text encoder} using CLIP vocabulary on 75\% train split.}
        \setlength{\tabcolsep}{3pt}
        \label{tab:text}
\begin{tabular}{@{}c|c|cc@{}}
\toprule
\multirow{2}{*}{Text encoder} & \multirow{2}{*}{Fine-tune} & \multicolumn{2}{c}{mAP}        \\ \cmidrule(l){3-4} 
                              &                        & \multicolumn{1}{c|}{0.5} & Avg \\ \midrule
No encoder                & -                      &  \multicolumn{1}{c|}{6.9}  & 11.5 \\ \midrule
CLIP-text                   & \xmark                    & \multicolumn{1}{c|}{36.9}  & 22.8  \\
CLIP-text                   & \cmark                      & \multicolumn{1}{c|}{38.2}  & 24.9  \\ \bottomrule
\end{tabular}
    \end{minipage}%
    \hfill
    \begin{minipage}{.45\linewidth}
      \centering
\renewcommand{\tabcolsep}{10pt}
\caption{Transformer vs. CNN on ActivityNet under $75\%$ seen label setting.}
\label{tab:emb}
\begin{tabular}{l|c|c}
\toprule
\multirow{2}{*}{\textbf{Network}} &
  \multicolumn{1}{c}{\textbf{mAP}} \\ 
  \cmidrule(l){2-3}
& \textbf{0.5}  & \textbf{Avg}  \\ \midrule
1D CNN  & 29.0          & 19.3          \\
MS-TCN  & 33.5          & 21.4          \\
Transformer & \textbf{38.2} & \textbf{24.9} \\
\bottomrule
\end{tabular}
    \end{minipage} 
\end{table}

\noindent\textbf{Vision to language cross-adaptation}
\textcolor{black}{
To further demonstrate the effects of cross-modal adaptation, we performed detailed ablation with the CLIP pre-trained variant of {\shortmodelname} and the results are shown in Table \ref{tab:crossmod}. To better leverage language priors, we used a transformer encoder \cite{vaswani2017attention} to condition the language embedding $F_{lang}$ using the masked foreground video representation $F^{fg}_{vis}$. From Table \ref{tab:crossmod}, we witness that there is atleast $1.4 \%$ of gain in avg mAP with the cross-modal adaptation. We observe that this gain is also achieved with negligible gain in compute. Therefore, we choose the cross-modal adaptation in all the rest of our experiments.}

\begin{table}[]
\centering
\renewcommand{\tabcolsep}{8pt}
\caption{We demonstrate that performing post model vision-to-language cross-modal adaptation can yield the better performance with fewer extra FLOPs and parameters.}
\label{tab:crossmod}
\begin{tabular}{@{}ccccc@{}}
\toprule
\multirow{2}{*}{Pre-Train} &
  \multirow{2}{*}{\begin{tabular}[c]{@{}c@{}}Language\\ Prompt\end{tabular}} &
  \multirow{2}{*}{V $\rightarrow$ L} &
  \multirow{2}{*}{avg mAP} &
  \multirow{2}{*}{\begin{tabular}[c]{@{}c@{}}Params\\ (M)\end{tabular}} \\
     &     &     &    &    \\ \midrule
CLIP & \cmark & \xmark  & 23.5 & 46.5 \\
CLIP & \cmark & \cmark  & 24.9 & 50.2 \\ \bottomrule
\end{tabular}
\end{table}

\noindent\textbf{Textual context length}
\textcolor{black}{
Since we are dealing with learnable context, one natural question is that:
\textit{How many context tokens should
be used? And is it better to have more context tokens?}. 
We study this factor on ActivityNet dataset. 
The results in
Table \ref{tab:numtok} suggest that having longer context length for TAD task benefits. It indicates that
having more context tokens leads to better performance
and that positioning the class token in the middle gains
more momentum with longer context length. Hence we fix $50$ as context length for our experiments. We observe that a few words are somewhat relevant to the tasks, such as $``playing"$, $``swimming"$ for ActivityNet. But when connecting all the nearest words together, the prompts do not make much sense as seen in \cite{zhou2021learning}. We conjecture that the learned vectors might encode meanings that are beyond the existing vocabulary. However, for video-understanding tasks having larger learnable context token benefits. }

\begin{table}[]
\renewcommand{\tabcolsep}{16pt}
\centering
\caption{Analysis of no of tokens for text context on 75\% seen split on ActivityNet}
\label{tab:numtok}
\begin{tabular}{@{}c|cc@{}}
\toprule
\multirow{2}{*}{Number of Context Token} & \multicolumn{2}{c}{mAP}          \\ \cmidrule(l){2-3} 
                                         & \multicolumn{1}{c|}{0.5}  & Avg  \\ \midrule
10                                       & \multicolumn{1}{c|}{36.3} & 24.2 \\
30                                       & \multicolumn{1}{c|}{37.6} & 24.7 \\
\textbf{50}                                       & \multicolumn{1}{c|}{\textbf{38.2}} & \textbf{24.9} \\
70                                       & \multicolumn{1}{c|}{38.0} & 24.8 \\ \bottomrule
\end{tabular}
\end{table}

\section{Conclusion}

In this work, we have proposed a novel {\em \modelname} (\shortmodelname) model
for the under-studied yet practically useful zero-shot temporal action detection (ZS-TAD).
It is characterized by
a {\em parallel}
localization (mask generation) and classification architecture
designed to solve the localization error propagation problem
with conventional ZS-TAD models.
For improved optimization, we further introduced an inter-branch consistency regularization to exploit their structural relationships.
Extensive experiments on ActivityNet and THUMOS have demonstrated that our \shortmodelname{} yields state-of-the-art performance under both zero-shot and supervised learning settings.

%
%
\bibliographystyle{splncs04}
\bibliography{egbib}
\end{document}